\documentclass[10pt,twocolumn,letterpaper]{article}

\usepackage[pagenumbers]{cvpr} 

\usepackage{graphicx}
\usepackage{amsmath}
\usepackage{amssymb}
\usepackage{booktabs}
\usepackage{algorithm}
\usepackage{algorithmic}
\usepackage{multirow}

\usepackage[pagebackref,breaklinks,colorlinks]{hyperref}
\usepackage{fontawesome}

\usepackage[capitalize]{cleveref}
\crefname{section}{Sec.}{Secs.}
\Crefname{section}{Section}{Sections}
\Crefname{table}{Table}{Tables}
\crefname{table}{Table }{Tabs.}

\def\cvprPaperID{1730} 
\def\confName{CVPR}
\def\confYear{2022}

\begin{document}

\title{DVCFlow: Modeling Information Flow Towards Human-like Video Captioning}

\author{
Xu Yan$^{1,2}$, Zhengcong Fei$^{1,2}$, Shuhui Wang$^{1, *}$, Qingming Huang$^{1,2,3}$, Qi Tian$^{4}$\\
	$^1$ Key Lab of Intell. Info. Process., Inst. of Comput. Tech., CAS, Beijing, China\\
	$^2$ University of Chinese Academy of Sciences, Beijing, China\\
	$^3$ Peng Cheng Laboratory, Shenzhen, China\\
	$^4$ Cloud BU,  Huawei Technologies, Shenzhen, China\\
	\tt\small{\{yanxu19s, feizhengcong, wangshuhui\}@ict.ac.cn,} \\ \tt\small{qmhuang@ucas.ac.cn, tian.qi1@huawei.com}
}

\maketitle

\begin{abstract}
Dense video captioning (DVC) aims to generate multi-sentence descriptions to elucidate the multiple events in the video, which is challenging and demands visual consistency, discoursal coherence, and linguistic diversity. 
Existing methods mainly generate captions from individual video segments, lacking adaptation to the global visual context and progressive alignment between the fast-evolved visual content and textual descriptions, which results in redundant and spliced descriptions.
In this paper, we introduce the concept of information flow to model the progressive information changing across video sequence and captions. 
By designing a Cross-modal Information Flow Alignment mechanism, the visual and textual information flows are captured and aligned, which endows the captioning process with richer context and dynamics on event/topic evolution.
Based on the Cross-modal Information Flow Alignment module, we further put forward DVCFlow framework, which consists of a Global-local Visual Encoder to capture both global features and local features for each video segment, and a pre-trained Caption Generator to produce captions. 
Extensive experiments on the popular ActivityNet Captions and YouCookII datasets demonstrate that our method significantly outperforms competitive baselines, and generates more human-like text according to subject and objective tests.
\end{abstract}


\begin{figure}
	\centering
	\includegraphics[width=0.9\columnwidth]{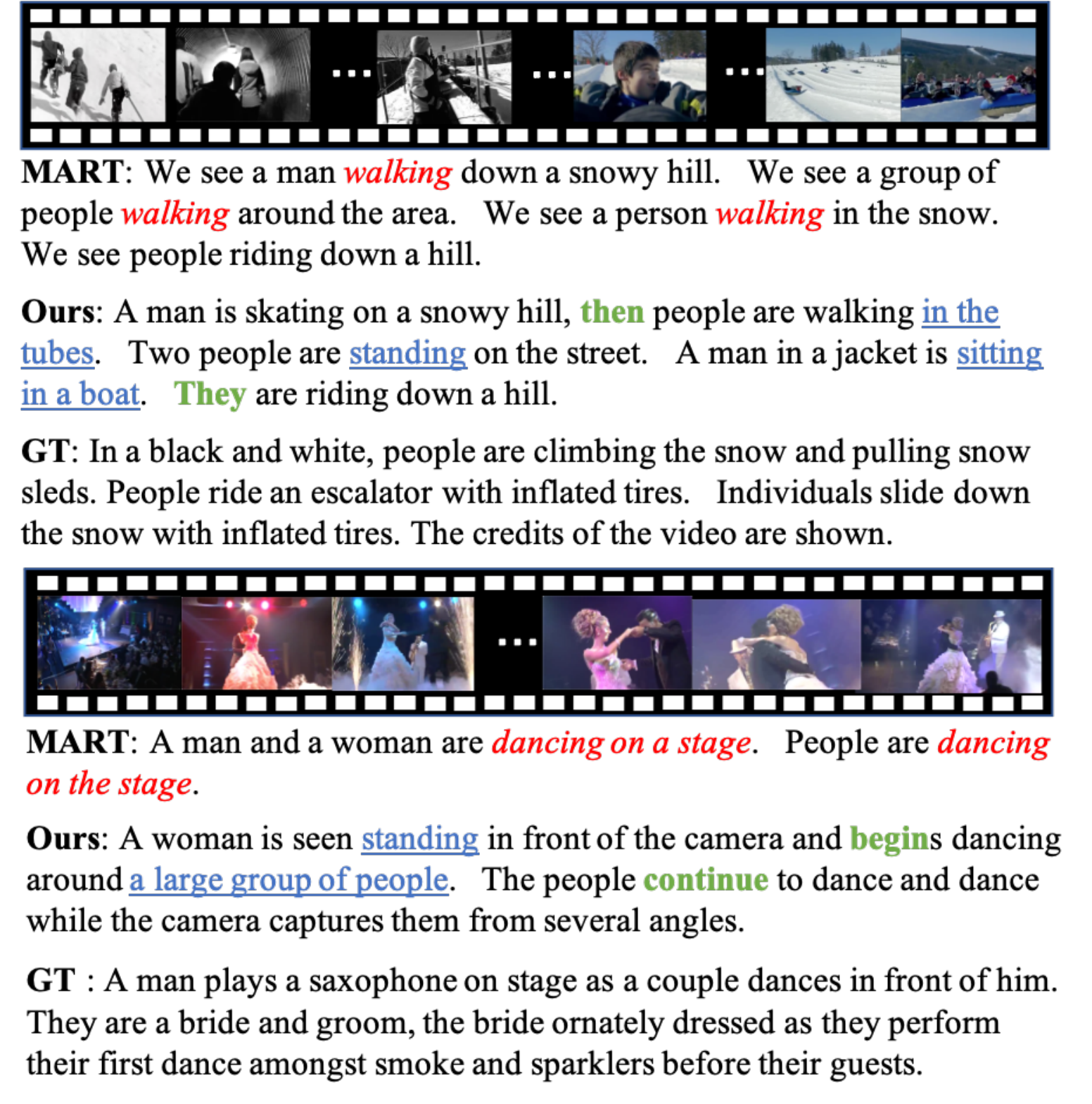}
	\caption{Comparison between the state-of-the-art video description approach MART~\cite{lei2020mart} and our proposed DVCFlow. Our approach generates more visually consistent and semantically coherent descriptions with less redundancy. Videos are from ActivityNet Captions~\cite{caba2015activitynet,krishna2017dense}; red/italic indicates repeated patterns, blue/underscore indicates visually consistent phrases, and green/bold highlights the coherence of the text.}
	\label{fig:0}
\end{figure}

\section{Introduction}

Video has become the dominant form of information delivery on the Internet, as it delivers richer information than pure images or text. However, most videos do not have corresponding descriptions. Therefore, dense video captioning is an essential task to benefit video content understanding, retrieval, and sharing. Despite the widespread availability of video captioning at the sentence level~\cite{vinyals2015show,donahue2015long,yao2015describing,chen2018less,wang2018video,zhang2021open}, how to automatically generate fine-grained paragraphs~\cite{yu2016video,wang2018bidirectional,wang2020event} is need to be explored more comprehensively. 
Dense video captioning~\cite{krishna2017dense} is one of the potential research directions related to cross-media understanding and vision-language, which aims to recognize multiple meaningful segments in the long video and then describe these events by generating human-like textual descriptions.

Several methods have been proposed for dense video captioning. These methods mainly adopt a two-stage pipeline, where the systems first identify candidate events from the original videos and then generate multi-sentence paragraph descriptions in a progressive manner~\cite{donahue2015long,yao2015describing, venugopalan2015sequence,krishna2017dense,wang2018bidirectional,xiong2018move}. 
It has been stressed by previous work ~\cite{park2019adversarial,lei2020mart} that generating descriptions for videos with the two-stage framework is very challenging due to inaccurate event proposals. 
Therefore, the task is simplified as generating multi-sentence descriptions for the given list of segments, removing the event proposal process and focusing on decoding better captions. 
In this work, we follow their settings to focus on generating more human-like captions. 

As shown in \Cref{fig:0}, there are still many problems for the existing methods, such as repetitive sub-sentences, visual inconsistent phrases, and incoherent captions.
These problems are mainly caused by the limitation on modeling the relationship among video event segments and the misalignment between video event segments and multi-sentence descriptions.
For the first issue, several existing methods ~\cite{krishna2017dense,xiong2018move,zhou2018end,park2019adversarial} only consider the current video event segment and ignore the relationship among events. This lack of global visual context limits the understanding of the current video event. 
Lei {\it et al.}~\cite{lei2020mart} point out this problem and focus on historical visual information, but still discard the exploration of future events. 
For the second issue, the misalignment is mainly reflected in the repetitive and vague phrases among the captions, which misses the specific details of the video event segments.
This is because almost all methods directly establish the translation from the video event segments to the corresponding captions, without modeling the impact of the event changing across video segments on the topic evolution of the caption sequence.


To tackle aforementioned issues, we aim to generate human-like paragraph descriptions.
We consider that each video event segment can bring new information compared to the past video content. 
Correspondingly, the caption of the video segment should focus on the new information rather than the general content. 
Therefore, the progressive information changing across video event segments and multi-sentence captions should be modeled as a multimodal \textbf{information flow}, which is inspired by the human cognitive process that the historical and future information is beneficial for understanding the current video event segment. 

Accordingly, we propose DVCFlow model, which consists of three modules, {\it i.e.}, Global-local Visual Encoder, Cross-modal Information Flow Alignment Module, and Caption Generator.
First, we employ the off-the-shelf bidirectional Transformer layers~\cite{vaswani2017attention} as the Global-local Visual Encoder, which encodes the video event segments in a multi-level fashion.  
The Cross-modal Information Flow Alignment Module aims to model and align the visual and textual information flows. To model the visual information flow, we design a uni-directional visual flow module to capture the information changing across video event segments. 
Considering the limited size of the dataset and the need for long text understanding for dense video captioning, we fine-tune the large-scale pre-trained language model as the textual representation that better captures the textual information flow across the captions. Then we align the visual and textual information flows by minimizing the divergence between them. 
On this basis, the visual information changing is used as the guidance to generate more visually consistent and semantically coherent captions. 
Here, the captions are generated by the Caption Generator progressively based on the multimodal input representations and the visual information changing.

DVCFlow could better understand video event segments globally and well model the impact of visual information changing on the topic evolution of the caption sequence.
Therefore, our method significantly reduces sentence redundancy and generates human-like dense video captions.
We conduct experiments on the ActivityNet Captions and YouCookII datasets. 
Automatic and human evaluation results show that DVCFlow can generate more visually consistent, discoursal coherent, and textually informative descriptions compared to the competitive baseline methods. In addition, ablation studies demonstrate the effectiveness of modeling global visual representations and aligning the visual and textual information flows.

Our contributions are as follows: 
\begin{itemize}
    \item We propose a novel DVCFlow model to improve the visual consistency, discoursal coherence and textual diversity of dense video captioning. Specifically, our model encapsulates a global-local Transformer for visual representations and a fine-tuned language model for long-term text understanding.
    \item To exploit the cross-modal information, we introduce an information flow alignment mechanism, which sequentially models and aligns the intrinsic information changing across video events and text descriptions. 
    \item We demonstrate that the DVCFlow surpasses competitive baselines and achieves a new state-of-the-art on the ActivityNet Captions and YouCookII datasets. To improve reproducibility and foster future research, we will release the source code in the future. 
\end{itemize}

\section{Related Work}
\label{sec:related_work}

\subsection{Dense Video Captioning}

Most video captioning models\cite{vinyals2015show,donahue2015long,yao2015describing,chen2018less,wang2018video} generate single captions for trimmed short video clips with an average duration of 6-25 seconds. The video paragraph captioning task~\cite{yu2016video,rohrbach2014coherent,chen2021scan2cap} has also been proposed to provide multi-sentence descriptions with several segments. Afterward, Krishna \etal~\cite{krishna2017dense} introduced the task of dense-captioning events and proposed a large-scale benchmark named ActivityNet Captions. 

The common paradigm in modern dense video captioning usually adopts a two-stage framework. The framework candidates events through video action detection and then generates a coherent paragraph~\cite{donahue2015long,yao2015describing, venugopalan2015sequence,krishna2017dense,wang2018bidirectional,xiong2018move}.
The approaches can be mainly divided into two categories: enhancing the encoding ability and assisting the decoding process. On the one hand, Wang \etal~\cite{wang2018bidirectional} proposed a bidirectional sequence encoder to exploit both past and future contexts. Wang \etal~\cite{wang2020event} used a multi-level representation to enrich visual features. COOT~\cite{2020COOT} designed richer visual representations at different granularity levels.
On the other hand, MFT~\cite{xiong2018move} selected the key temporal segments to reduce the repetition issues. Wang \etal~\cite{wang2019controllable} and Hou \etal~\cite{hou2019joint} further extracted the Part-Of-Speech information to ensure the correctness of the grammar. In addition, automatic speech recognition tokens were merged as input~\cite{hessel2019case,iashin2020better,iashin2020multi,shi2019dense,lin2021vx2text} to guide the generation by learning rich information across video, sound, and text.
Recently, Transformer architecture~\cite{vaswani2017attention} is employed, where the multi-head self-attention could learn a global interaction among input objects in different subspaces~\cite{zhou2018end,lei2020mart,sun2019videobert,zhu2020actbert}.
M.Transformer~\cite{zhou2018end} proved that a Transformer-based model could achieve higher performance than the LSTM-based model~\cite{xiong2018move,zhou2019grounded,zhang2018cross}. In the next, a series of works are leveraged, \emph{e.g.}, VideoBERT~\cite{sun2019videobert}, Transformer-XL~\cite{dai2019transformer}, and MART~\cite{lei2020mart}. DVCFlow is also based on Transformer, and the used segment features have the entire video context to model relationships among all events.

\subsection{Cross-modal Alignment}

The research on image-text cross-modal alignment has drawn intensive attention. Specifically, VisualBERT~\cite{2019VisualBERT} directly migrates BERT to vision-language field. Unicoder-VL~\cite{li2020unicoder} and VL-BERT~\cite{su2019vl} further enhance the region-level vision-language alignment. Next, ViLBERT~\cite{lu2019vilbert} and LXMERT~\cite{tan2019lxmert} both utilize a more detailed two-stream encoder structure. Furthermore, VideoBERT~\cite{sun2019videobert} and ActBERT~\cite{zhu2020actbert} use a BERT-type objective to build a bridge between video and text.
In short, our approach belongs to the Transformer-based encoder-decoder framework, where the decoder adopts a pre-trained language model. The novelty lies in the exploration of the sequential change in the semantic information of the captions, which has not been previously explored.

\section{Approach}
\label{sec:approach}

Focusing on human-like dense video captioning, we directly utilize the annotated video event segments in the ground-truth. Before introducing our method in detail, we first define some terms. Given a long video $v={\{v_1, v_2, \cdots, v_T\}}$, containing $T$ frames, which is divided into $K$ temporally ordered event segments $\{s_1, s_2, \cdots, s_K\}$, the corresponding captions are denoted as $\{c_1, c_2,\cdots, c_K\}$.
More specifically, the semantic information of the $i$-th caption $c_i$ is consistent with the visual feature of the $i$-th video segment $s_i$, where $i = 1, 2, \cdots, K$.

\begin{figure*}
	\centering
	\includegraphics[width=1.90\columnwidth]{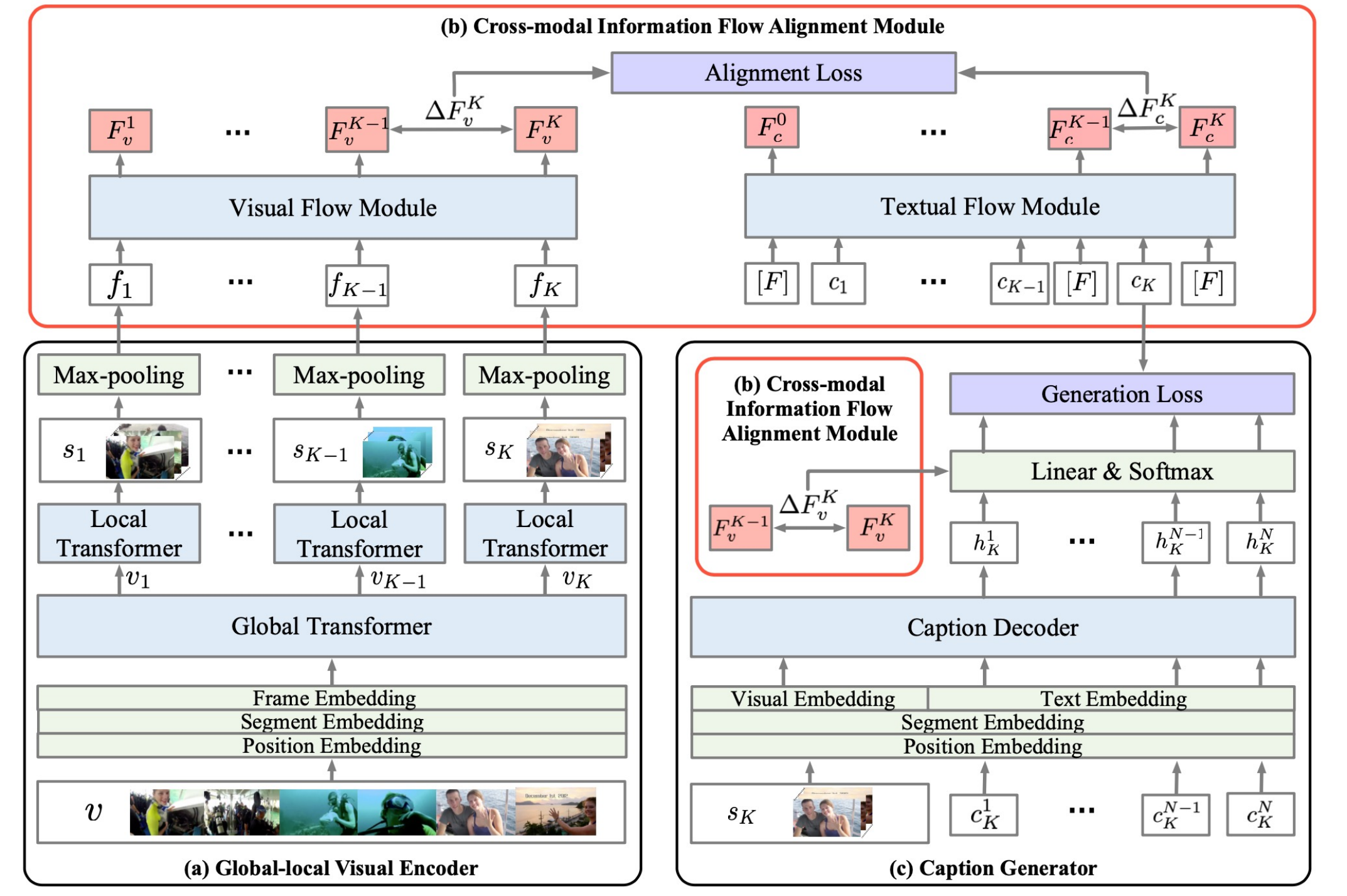}
	\caption{Overview of our proposed DVCFlow model. 
	(a) Global-local Visual Encoder contains a global transformer layer and a local transformer layer, which are both the off-the-shelf transformer layers with multi-head attention. 
	(b) Cross-modal Information Flow Alignment Module includes the Visual Flow Module and the Textual Flow Module. Visual Flow Module is a uni-directional transformer layer with masked multi-head attention for capturing the incremental visual influence $\Delta F^K_v$ (taking the last segment as an example). Textual Flow Module is the pre-trained language model for capturing the incremental textual influence $\Delta F^K_c$. Besides, we design an alignment loss to bridge the gap between the corresponding incremental visual and textual influences ($\Delta F^K_v$ and $\Delta F^K_c$).
	(c) Caption Generator is implemented with the pre-trained language model sharing weight with the textual flow module, which generates the caption of a video event segment based on the multimodal input representations and the incremental visual influence $\Delta F^K_v$.} 
	\label{fig:1}
\end{figure*}

\subsection{\textbf{DVCFlow Framework}}
\label{DVCFlow}
\subsubsection{\textbf{Model Overview}}
As shown in \Cref{fig:1}, DVCFlow is based on the encoder-decoder framework, which consists of three modules: 

($a$) The Global-local Visual Encoder aims to obtain the representations of local video event segments with global context. We employ the Transformer architecture for the visual encoder instead of LSTM since Xiong {\it et al.}~\cite{xiong2018move} proved that the Transformer-based models hold better performance and speed. 

($b$) The Cross-modal Information Flow Alignment Module is designed to capture and align the visual information flow and textual information flow. We devise a uni-directional visual flow module to model the impact of visual information changing, referred to as incremental visual influence $\Delta F_v^K$. For textual information flow, we fine-tune the pre-trained language model to model the semantic changing brought about by the current caption, referred to as incremental textual influence $\Delta F_c^K$.
Finally, we design an alignment loss to align the two incremental influences based on the multimodal information flow.

($c$) The Caption Generator is utilized to generate the caption based on the global-local visual features and the aligned visual influence of the current video segment, which is implemented with a language model sharing weights with the one in the textual flow module.

\subsubsection{\textbf{Global-local Visual Encoder}}
For the input of the visual encoder, we directly utilize the extracted video features following previous works~\cite{xiong2018move,lei2020mart} for a fair comparison with state-of-the-art methods. Specifically, we incorporate a segment embedding layer into the embedding layers to mark the timestamps of different segments in the entire video, which serves as proposal-free processing.

All the embedded video features are fed into a Global Transformer. The representation of the $i$-th video event segment can be extracted from the output of the Global Transformer with annotated timestamps, denoted as $s_i$, which holds a global view of the entire video, \emph{i.e.}, obtaining the historical and future information: 

\begin{equation}
    \{s_i\}^K_{i=1} = \text{Global-Transformer}(v),
    \label{s_i}
\end{equation}
where the Global-Transformer is implemented by a bidirectional Transformer, and  $K$ denotes the number of segments/captions in the given long video $v$. 
Among the segments, future information is essential, especially when generating the first several captions of a paragraph.

However, only using the global Transformer is difficult to express the more specific content from the current segment. To tackle this, we further develop a Local-Transformer to enhance the specific video event segment representations. Correspondingly, the \Cref{s_i} is modified as:
\begin{equation}
\begin{aligned}
    \{v_i\}^K_{i=1} = \text{Global-Transformer}(v),\\
    s_i= \text{Local-Transformer}(v_i).
\end{aligned}
\end{equation}
Note that the Local-Transformer has the same structure as Global-Transformer. 
Then we adopt a max-pooling layer to reduce redundant information so that we can obtain key representation $f_i$, which is further fed into visual flow module:
\begin{equation}
\label{f_i}
     f_i = \text{Max-pooling}(s_i).\\
\end{equation}

\subsubsection{\textbf{Cross-modal Information Flow Alignment}}
Intuitively, when the video events among adjacent segments change, the differences in visual content will guide the topic evolution of the caption sequence. Therefore, aligning the visual content changes with the caption topic evolution is beneficial for generating visually consistent and semantically coherent descriptions. Therefore, we model visual and textual information flows to calculate the changes in segment and caption, respectively.

For modeling the visual information flow, we design the Visual Flow Module utilizing a unidirectional Transformer layer with the masked multi-head attention.
We calculate the $i$-th visual information flow $F^i_v$ as the following:
\begin{equation}
     F^i_v = \text{Visual-Flow}(f_1, f_2, \cdots, f_i),
\end{equation}
where $f_i$ denotes the $i$-th key representation among frames from \Cref{f_i}.

Thus, we utilize $\{F^1_v,\cdots,F^{K-1}_v, F^K_v\}$ to record the dynamic visual information flow across $K$ different segments, where the $K$-th flow $F^K_v$ is based on all the previous historical key representations $\{f_1, f_2, \cdots , f_K\}$. 
Then, we define the difference between the new visual information flow $F^i_v$ at the $i$-th video event segment and the previous information flow $F^{i-1}_v$ at the $(i-1)$-th segment as the incremental visual influence $\Delta F^i_v$ of the $i$-th segment, which can be formulated as:
\begin{equation}
\label{deltaFv}
    \Delta F^i_v = F^i_v - F^{i-1}_v.
\end{equation}

For the textual information flow, considering that there are lots of time spans in the untrimmed videos, the corresponding paragraph description is very long and the understanding procedure can be regarded as a long-text modeling task.
However, it is difficult to train a caption encoder and decoder from scratch for multi-sentence long text understanding due to the insufficient data size in the dataset, especially the rare but important words. To tackle this issue, we fine-tune the pre-trained language model GPT-2~\cite{radford2019language} as the textual flow module. 

The input sequence of the textual flow module can be denoted as $\{[F], c_1, \cdots, [F], c_K, [F]\}$, where $[F]$ is a special token representing the caption interval. In other words, the $i$-th textual information flow $F^i_c$ is the output of the textual flow module corresponding to the $i$-th special token $[F]$ in the input. The $i$-th textual information flow $F^i_c$ contains the history text information, same as visual information flow $F^i_v$:

\begin{equation}
    F^i_c = \text{Textual-Flow}(\{[F],c_1,\cdots,[F],c_i,[F]\}).
\end{equation}

So the incremental textual influence $\Delta F^i_c$ of the $i$-th caption can be formulated as:
\begin{equation}
\label{deltaFc}
    \Delta F^i_c = F^i_c - F^{i-1}_c.
\end{equation}

\subsubsection{\textbf{Caption Generator}}

For caption generator, we fine-tune the language model as the decoder, which shares weight with the one in the Cross-modal Information Flow Alignment Module. It can generate high-quality captions conditioned on the multimodal input representation and the aligned incremental visual influence. Thus, we can shift the attention of the generated caption according to the specific video segment during the paragraph generation process.

The Caption Generator takes input from both the $i$-th processed segment representation $s_i$ and the corresponding $i$-th text caption $\{c^1_i,\cdots,c^{N-1}_i, c^N_i\}$ in sequence, where $N$ is the length of the $i$-th caption and $1 \leq i \leq K $. 
Note that in the Caption Generator, the visual part utilizes the multi-head attention, while the caption part employs the masked multi-head attention. And we further add a trainable segment embedding layer to indicate whether the input token is from video or text, similar to BERT~\cite{devlin2018bert}:
\begin{equation}
    \begin{aligned}
    & E_v = \text{MHA}(s_i),\\
    & E_c = \text{Masked-MHA}(\{c^j_i\}^N_{j=1}),\\
    & \{h^j_i\}^N_{j=1} = \text{GPT-2}(\text{Embedding}([E_v;E_c])),
    \end{aligned}
\end{equation}
where $[;]$ denotes concatenation operation, $1 \leq j \leq N$, $\text{MHA}(\cdot)$ and $\text{Masked-MHA}(\cdot)$ correspond to multi-head attention and masked multi-head attention, respectively.

In contrast to the previous approaches, we additionally concatenate the hidden representations of the decoder $\{h^1_i, \cdots, h^{N-1}_i, h^N_i\}$ with the incremental visual influence $\Delta F^i_v$ at each time-step and feed them to an output layer, which makes the model have the capacity to fuse the visual and textual representations. The fusion procedure can be summarized as below:
\begin{equation}
    \begin{aligned}
    & H_i = [\Delta F^i_v ; \{h^j_i\}^N_{j=1}],\\
    & c_i = \{c^j_i\}^N_{j=1},\\
    & p(c_i|H_i) = \text{Softmax}(W^cH_i+b^c),\\
    \end{aligned}
\end{equation}
where [;] denotes concatenation operation, $W^c$ and $b^c$ denote trainable weights and bias.

\subsection{\textbf{Training Objectives}}
\label{lm}

The main target of DVCFlow is to generate a human-like multi-sentence description for the given video by exploiting the training procedure to optimize model parameters. 
In this work, we design two training objectives to optimize the model parameters, \ie, 1) Generation loss and 2) Alignment loss. 
Specifically, same as the conventional approaches, for generation loss, we calculate the Negative Log Likelihood (NLL) loss to ensure the semantic coherence and correctness of caption. Besides, as shown in \Cref{fig:1}, we model the mutual influence between captions and the video event segments. To obtain visually consistent and semantically coherent captions, we also optimize the model by minimizing the Mean Square Error (MSE) loss between visual information flow and textual information flow.

\subsubsection{\textbf{Generation Loss}}

Our DVCFlow model generates the human-like captions $c_K$ of the video event segment $v_K$ based on the global-local video event segment representation $s_K$ and the incremental visual influence $\Delta F_v^K$. Therefore, the generation loss could be formulated as follows:
\begin{equation}
\begin{aligned}
    \mathcal{L}_{NLL} &= -\sum_{i=1}^{K}\text{log} \prod_{j=1}^{N} p(c_i^{j}|c_i^{\textless j}, s_i, \Delta F_v^i)\\
    &= -\sum_{i=1}^{K}\text{log} \prod_{j=1}^{N} \mathcal{F}_{H_i^j},
\end{aligned}
\end{equation}
where $\mathcal{F}_{H_i^j}$ denotes the estimated probability of the $j$-th token in the $i$-th caption. The projection function $\mathcal{F}$ is used to reflect the hidden states to the probability of tokens:
\begin{equation}
    \mathcal{F} = \text{Softmax}(W^c [\Delta F_v^i; h_i^j]+b^c) \in \mathbf{R}^{|V|}, 
\end{equation}
where $h_i^j$ denotes the hidden states, $|V|$ refers to the vocabulary size, $W^c$ and $b^c$ are learnable parameters. 

\subsubsection{\textbf{Alignment Loss}}
To bridge the gap between visual information flow and textual information flow, we design the Alignment loss, \ie, Mean Square Error (MSE) loss, which aims to induce the model to understand the inner relationships between incremental visual influence and incremental textual influence. Here, the incremental visual influence $\Delta F^i_v$  among video event segments is obtained by \Cref{deltaFv}, while the incremental textual influence $\Delta F^i_c$ is calculated by \Cref{deltaFc}. Thus, MSE loss is used to minimize the distance between the two incremental influences, \ie, $\Delta F^i_v$ and $\Delta F^i_c$:
\begin{equation}
    \mathcal{L}_{MSE}= \sum_{i=1}^{K}||\Delta F^i_v -\Delta F^i_c||^2_2.
\end{equation}

The overall training objective of DVCFlow could be computed as follows:
\begin{equation}
    \mathcal{L} = \mathcal{L}_{NLL} + \lambda \mathcal{L}_{MSE}.
\end{equation}

\begin{table*}
   \begin{center}
   {\caption{Performance comparisons with different evaluation metrics on the ActivityNet Captions \textbf{test} split and YouCookII val split. The best score of all video dense captioning models is in bold font. All values are reported as a percentage (\%). Top scores are highlighted.}
     \label{tab1}}
   \begin{tabular}{lcccccccc}
     \toprule
      \multirow{2}{*} { Model }
      &\multicolumn{4}{c}{\textbf{ActivityNet Captions(\textbf{test})}} &\multicolumn{4}{c}{\textbf{YouCookII(\textbf{val})}} \\
      \cmidrule(r){2-5}\cmidrule(r){6-9}&BLEU@4 &METEOR &CIDEr &R@4$\downarrow$ &BLEU@4 &METEOR &CIDEr &R@4$\downarrow$\\
     \midrule 
     \multicolumn{9}{l}{\textbf{LSTM-based methods}} \\
     MFT~\cite{xiong2018move} &10.33 &15.09 &19.56 &15.88 &- &- &- &- \\
     HSE~\cite{zhang2018cross} &9.84  &13.78 &18.78 &13.22 &- &- &- &- \\ 
     \midrule
     \multicolumn{9}{l}{\textbf{LSTM-based methods with detection feature}} \\ 
     GVD~\cite{zhou2019grounded} &\textbf{10.70} &16.10 &22.20 &8.76 &- &- &- &- \\
       AdvInf~\cite{park2019adversarial}&9.91 &16.48 &20.6 &5.76 &- &- &- &-\\
     \midrule
     \multicolumn{9}{l}{\textbf{Transformer-based methods}} \\ 
       M.Transformer~\cite{zhou2018end} &9.75 &15.56 &22.16 &7.79 &7.62 &15.65 &32.26 &7.83\\
       Transformer-XL~\cite{dai2019transformer} &10.25 &14.91 &21.71 &8.79 &6.56 &14.76 &26.35 &6.3 \\
       MART~\cite{lei2020mart} &9.78 &15.57 &22.16 &5.44 &\textbf{8} &15.9 &\textbf{35.74} &4.39\\
     DVCFlow(Ours) &10.21 &\textbf{17.38} &\textbf{23.66} &\textbf{0.23} &7.87 &\textbf{16.5} &34.7 &\textbf{2.3}\\
     \bottomrule
   \end{tabular}
   \end{center}
  \end{table*}

\begin{table}
	\begin{center}
		\caption{Ablation study on ActivityNet Captions \textbf{val} split. “- flow alignment” denotes a variant without MSE Loss. We report BLEU@4(B@4), METEOR(M), CIDEr(C) ,and R@4.}
		\label{ablation}
		\begin{tabular}{lcccc}
			\toprule
			Model &B@4  &M &C &R@4↓\\
			\midrule    
			DVCFlow &10.52 &17.31 &24.32 &0.20 \\
			- flow alignment($\mathcal{L}_{MSE}$) &9.11 &16.72 &20.11 &0.27 \\ 
			- global features &10.23 &16.01 &23.56 &0.23 \\
			- pre-trained weights &9.62 &16.54 &22.18 &0.29 \\
			\bottomrule
		\end{tabular}
	\end{center}
\end{table}

\section{Experiments}
\label{sec:experiments}

\subsection{Experimental Settings}

\subsubsection{\textbf{Datasets.}}

The ActivityNet Captions dataset~\cite{krishna2017dense} built on ActivityNet v1.3~\cite{caba2015activitynet} contains 10,009 videos for training, and 4,917 for validation. The videos in this dataset are 3 minutes long on average and longer than Microsoft Research Video Description Corpus (MSVD~\cite{chen2011collecting}) and Microsoft Research Video to Text (MSR-VTT~\cite{xu2016msr}) for general video captioning. 
On average, each set of annotations contains 3.65 sentences. We use splits provided by GVD~\cite{zhou2019grounded}, where the validation is split into two subsets: 2,459 videos for validation and 2,457 videos for test, which ensure the test videos will not be seen in the validation set.
YouCookII~\cite{zhou2018towards} contains 1,333 training videos and 457 validation videos. 
On average, there are 7.7 segments for each video in this dataset.

\subsubsection{\textbf{Evaluation Metrics.}}
The main evaluation metrics of dense video captioning include BLEU@4~\cite{papineni2002bleu},  METEOR~\cite{banerjee2005meteor}, CIDEr~\cite{vedantam2015cider}, and R@4~\cite{xiong2018move}. 
Among these metrics, BLEU@4 and CIDEr are evaluation metrics based on $n$-gram matching, which calculate the usage accuracy of words and measure the overall sentence fluency. In comparison, METEOR is a metric based on embedding representations, which could calculate the semantic distance between synonyms and reflect the degree of semantic correctness. Therefore, it has a higher correlation with human judgment. R@4 evaluates the degree of repetition between ground-truth and generated text. In our opinion, to some extent, if R@4 becomes lower (the lower, the better), it will inevitably lead to a decrease in the performance of CIDEr and BLEU@4. Therefore, in order to evaluate the performance of dense video captioning more comprehensively, we advise the human evaluation, and focus on R@4 and METEOR, instead of pursuing the only improvement of CIDEr and BLEU@4.

\subsubsection{\textbf{Implementation Details.}}

For video features, we use aligned appearance and optical flow features sampled at 2FPS, provided by M.Transformer~\cite{zhou2018end}. We truncate the video sequence longer than 900 frames. As for the text, we use the tokenizer of GPT-2~\cite{radford2019language} to tokenize the text to tokens. In contrast to the prevalent approaches of discarding words that appeared less than 3 or 5 times, it is available with a larger vocabulary, which helps to produce more informative tokens. We set the hidden size to 768. The number of attention heads is set to 12. Specifically, we use AdamW optimizer with an initial learning rate of 1e-4, $L2$ weight decay of 0.01, and linear learning rate warmup over the first 5 epochs. During the training process, the alignment loss was going down with the generation loss and then tended to be stable. For decoding strategy, we use top-$k$ sampling as we find that top-$k$ sampling can generate more informative and diverse captions while greedy search always leads to general, repetitive sentences. $k$ is set to 10. During the experiment, it costs about 24 hours on 4 GeForce RTX 3090 GPUs to train the model, and the inference time is about 1 hour on 1 GeForce RTX 3090 GPU on ActivityNet Captions dataset. Note that we used mixed precision for training.

\subsection{Quantitative Analysis}  
\Cref{tab1} shows the performances of different models. We can make some observations as bellows:

(1) Overall, the results across four evaluation metrics consistently indicate that our proposed DVCFlow achieves superior performances against other state-of-the-art techniques. In particular, the METEOR score, which can be considered the most valuable metric, can achieve 17.38 and 16.5 on two datasets. The reason why the R@4 score is lower than ground truth lies in that we filter out the captions with repetitive phrases in the decoding procedure.

(2) Richer basic visual features are crucial for performance improvement, \emph{e.g.}, COOT~\cite{2020COOT}, OEG~\cite{zhang2020object}, and Sub-GC~\cite{zhong2020comprehensive}. While we fed the general features without extra information into our model, same as MART, proving the effectiveness of our framework by higher scores.

(3)The improvement is obvious on the ActivityNet Captions dataset but a bit gentle on YouCookII dataset. This is because the videos of YouCookII are about cooking, and there are few semantic transformations among events.

\subsection{Ablation Study}
In this section, we further discuss the effectiveness of the three parts in our framework.
First, we test the variant by removing the cross-modal information flow alignment module, which is designed to bridge the gap between video event segments and multi-sentence captions. 
Second, to explore the effectiveness of global features, we perform ablation studies, where we replace the global Transformer with an extra local Transformer to discard the global view while leaving the number of parameters unchanged.
Moreover, we design a variant, where we train the caption generator and textual flow module from scratch, to verify the impact of the pre-trained language model on the long text representation. The evaluation results are reported in \Cref{ablation}.

According to the results, all the three aforementioned techniques are effective (comparing Row 2-4 with Row 1). We observe that the alignment brings about a significant improvement. Besides, integrating the global features also improves the performance to some extent, which demonstrates that the global part is beneficial for understanding the video event segment. When removing the pre-trained weights, the score decreased significantly, indicating that the pre-trained language model is essential. When combing all three techniques, we find that they complement each other and see the highest gain on all metrics.

\begin{figure*}
    \centering
	\includegraphics[width=1.90\columnwidth]{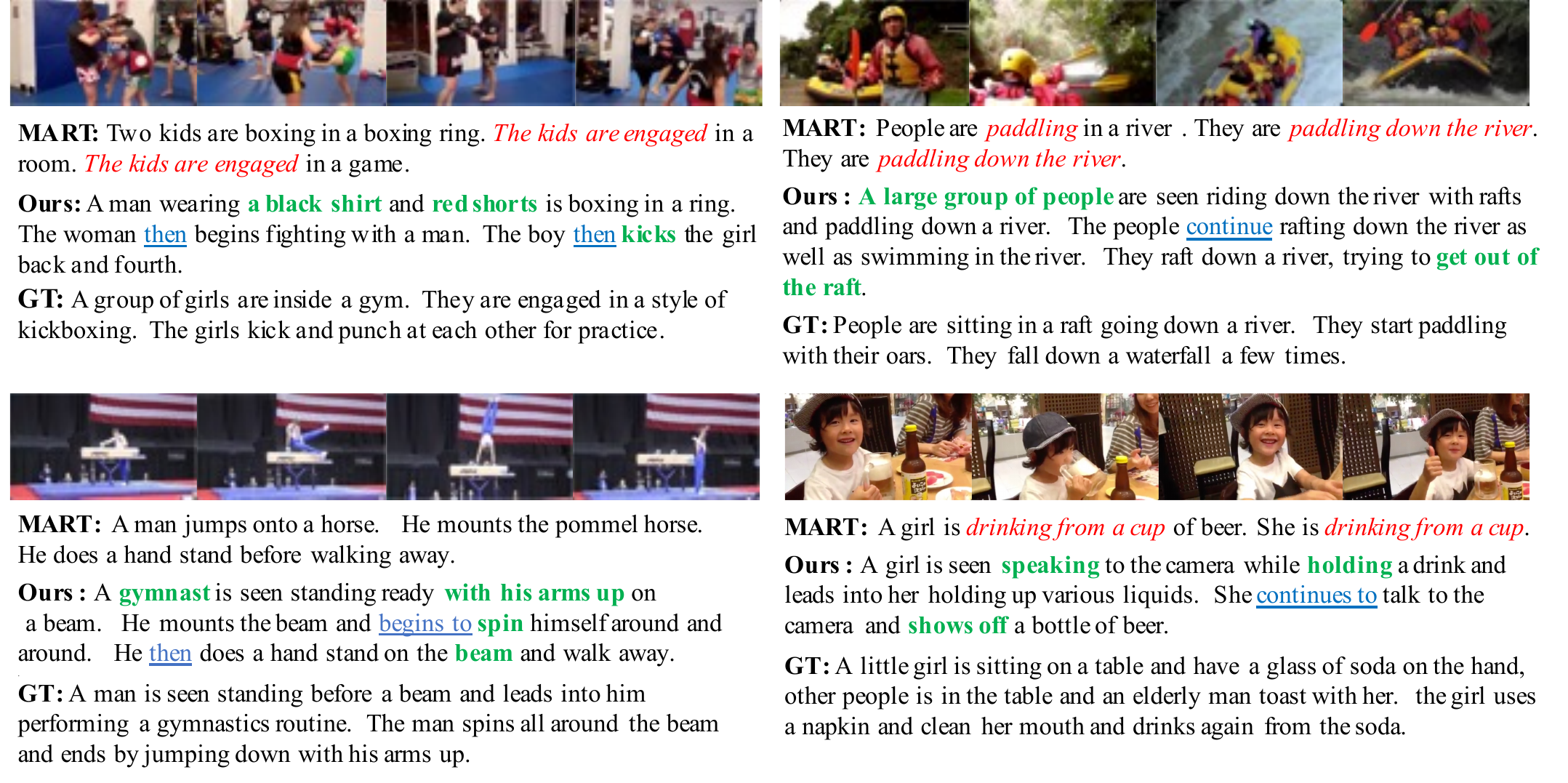}
	\caption{ Qualitative results on ActivityNet Captions. Colored words highlight relevant content to the segment. Our DVCFlow generates more informative and coherent captions as compared to the baseline.}
	\label{case}
\end{figure*}

\subsection{Qualitative Analysis}

\Cref{case} shows some examples with human-annotated ground-truth and descriptions generated by MART and our DVCFlow. From these examples, it is obvious that our model achieves a big step towards human-like dense video captioning since there are more conjunctions and fine-grained details instead of simple and general sentences.
For the first case, compared to the sentence segment "two kids" in the sentence generated by MART, "wearing a black shirt and red shorts" in our DVCFlow depicts more detailed and non-repeatable video content, which alleviates the problem of frequently repeated and general phrases in descriptions. 
In addition, DVCFlow could obtain more phrases regarding temporal information, such as "begin to" and "continue to", probably because our model has a global visual view, which effectively enhances the textual coherence. 
More visualization results can be found in the supplementary.

\subsection{Human Evaluation}
According to the results of automatic metrics, our method demonstrates the capability of generating high-quality paragraph descriptions for videos. Compared to those by other methods, the descriptions produced by our method are often more human-like.

To better understand how satisfactory the sentences generated from different methods are, we also conduct a human evaluation to compare our DVCFlow against state-of-the-art approach. 
Twelve well-educated evaluators are employed and a subset of 100 videos is randomly selected from ActivityNet Captions test split for the subjective evaluation. All evaluators and videos are randomly organized into four groups. We show each group both the sentences generated by different approaches and ask them: which caption is better or the two models are in a draw, with respect to \ie,  \emph{visual consistency, discoursal coherence,} and \emph{textual diversity}, respectively. 
Note that the models are anonymous, and the samples are shuffled. 
The results of the sample comparisons are obtained through a voting mechanism when each member of the group vote independently.  We calculate the scores of the human evaluation in \Cref{human}. We can obtain some observations from the results in the table. 
Overall, our DVCFlow is clearly the winner in terms of three different human-involved metrics. In particular, our DVCFlow has achieved a winning rate of more than 50\% in consistency and diversity, which indicates that our model holds the capability to effectively generate consistent as well as informative descriptions.

\begin{table}
    \begin{center}
    {\caption{Human evaluation on ActivityNet Captions test set w.r.t. consistency, coherence, and diversity. }
    \label{human}}
    \begin{tabular}{lccc}
    \toprule
    &Our wins(\%) &MART wins(\%)  &Ties(\%) \\
    \midrule 
    Consistency &62 &22 &16  \\
    Coherence &43 &28 &29 \\ 
    Diversity &56 &18 &26 \\ 
    \bottomrule
   \end{tabular}
  \end{center}
 \end{table}

\section{Limitation}
To solve the problem of generating multi-sentence texts for long videos, we propose a flow-based alignment mechanism. But the information flow requires that video clips have strong semantic relevance, which is suitable for natural long videos rather than synthetic videos. In addition, our model is the first attempt to solve the problem that sentence diversity will decrease when low frequency words are discarded in the pre-training stage, but it may lead to false associations. Details of bad cases are in the appendix.

\section{Conclusion}
\label{sec:conclusion}

In this paper, we propose a novel DVCFlow framework for informative, coherent and dense video captioning. We design an information flow mechanism to sequentially align the cross-modal context changing flow, based on a segment encoder with comprehensive global visual information and a pre-trained language decoder to enhance the language modeling capability for long text. Extensive experiments on two datasets, ActivityNet Captions and YouCookII, demonstrate that our proposed model achieves state-of-the-art performances and obtains more visually consistent and semantically coherent descriptions with less redundancy than advanced methods. 
In future works, towards more human-like descriptions, we are curious about extracting more effective basic video features, better utilization of the caption history, and solving the exposure bias issues.

{\small
\bibliographystyle{ieee_fullname}
\bibliography{sample-base}
}

\end{document}